\documentclass[10pt,twocolumn,letterpaper]{article}

\usepackage{cvpr}
\usepackage{times}
\usepackage{epsfig}
\usepackage{graphicx}
\usepackage{subfig}

\usepackage{amsmath}
\usepackage{amssymb}
\usepackage[numbers,sort&compress]{natbib}

\usepackage[pagebackref=true,breaklinks=true,letterpaper=true,colorlinks,bookmarks=false]{hyperref}

 \cvprfinalcopy % *** Uncomment this line for the final submission

\def\cvprPaperID{14} % *** Enter the CVPR Paper ID here

\ifcvprfinal\fi

\newif\ifsqueeze
\squeezetrue  % set \squeezetrue or \squeezefalse to enable or disable squeezing
\ifsqueeze

\else

\fi

\begin{document}

%%%%%%%%% TITLE
\title{Can poachers find animals from public camera trap images?}

\maketitle

\begin{abstract}
To protect the location of camera trap data containing sensitive, high-target species, many ecologists randomly obfuscate the latitude and longitude of the camera when publishing their data. For example, they may publish a random location within a 1km radius of the true camera location for each camera in their network. In this paper, we investigate the robustness of geo-obfuscation for maintaining camera trap location privacy, and show via a case study that a few simple, intuitive heuristics and publicly available satellite rasters can be used to reduce the area likely to contain the camera by 87\% (assuming random obfuscation within 1km), demonstrating that geo-obfuscation may be less effective than previously believed.
\end{abstract}

\section{Introduction} \label{sec:intro}
Monitoring biodiversity quantitatively can help us understand the connections between species decline and pollution, exploitation, urbanization, global warming, and conservation policy. Researchers study the effect of these factors on wild animal populations by monitoring changes in species diversity, population density, and behavioral patterns using camera traps. Camera traps are placed at specific, often hard-to-reach locations in the wild, and capture images when there is movement. Recently, there has been a large effort in the biology community to open-source camera trap data collections to facilitate reproducibility and provide verification (since mistakes can cause overestimates \cite{johansson2020identification}), as well as promote global-scale scientific analysis. By open-sourcing the images - not just metadata - collected across organizations, scientists studying a specific taxa can pool resources and leverage bycatch (images of species that were not the target of the original study, but are still scientifically valuable) from other camera trap networks. They will also be able to study animal behavior. A great deal of camera trap images are publicly available from all over the world, including via websites hosted by Microsoft AI for Earth and University of Wyoming \cite{lila} and Google \cite{wildlifeIn}. 

However, as mentioned on the Wildlife Insights FAQ page \cite{wildlifeInFAQ}, “Won’t Wildlife Insights images reveal the locations of endangered species to poachers?” They answer that “While Wildlife Insights is committed to open data sharing, we recognize that revealing the location for certain species may increase their risk of threat. To protect the location of sensitive species, Wildlife Insights will obfuscate, or blur, the location information of all deployments made available for public download\footnote{Public downloads are not yet available in Wildlife Insights} so that the exact location of a deployment containing sensitive species cannot be determined from the data. Practices to obfuscate the location information associated with sensitive species may be updated from time to time with feedback from the community.”  Community science (also known as citizen science) initiatives have also obfuscated locations to protect endangered species, such as eBird and iNaturalist, as there have been cases where community science and/or other open source data has informed poaching \cite{citizenScience}.

While obfuscating locations is encouraging, it is not clear whether it is sufficient to prevent geolocalization, or whether it is also necessary to blur or otherwise obfuscate portions of the images themselves. For example, for images in cities, geolocalization is typically based on recognizable landmarks and geometries, such as relationships between buildings and roads, heights of buildings, and strong architectural features or signage. The more recognizable a feature (such as a famous landmark or horizon), the easier an image is to geolocate. If you see an image containing the Chrysler building, it's easy to know that you're most likely in NYC. If you can see the outline of Mount Rainier, you're most likely in or near Seattle. An image of the exterior of a nondescript chain hotel or stretch of highway might be more difficult to place. We believe both human intuition and automated methods such as \cite{weyand2016planet} take advantage of these features.  The same might be said of camera trap imagery: if your image contains a noticeable landmark (for example a set of large rocky outcroppings or a body of water), it might be easier to estimate its location. In contrast, an image of dense undergrowth is only as potentially geolocalizable as your ability to recognize and model the distributions of its visible flora and fauna, and your ability to estimate a latitudinal band based on the timings of sunrise and sunset. In order to make these data publicly available, it is imperative that we understand whether these camera trap images can reveal locations, and if so, how to prevent locations from being revealed.

In this paper, we investigate how ``obfuscated'' these camera locations are, both with existing off-the-shelf geolocalization models and with a human-in-the-loop algorithmic approach we define as a proof-of-concept. We show that while existing models struggle to accurately locate camera trap images, a systematic method of filtering targeted to a specific conservation area using publicly available satellite rasters can be used to find specific candidate areas that are quite accurate, rendering the geo-obfuscation ineffective and indicating that the answer to our titular question is yes. %We point out camera placement strategies which may reduce the localizability of a camera location.

% , as well as a potential solution for blurring localizable features without removing animal observations. 
% \smb{since the cameras are somewhat static, I think we could try selecting a blur area and then using off-the shelf MegaDetector and DeepMask to unblur animal pixels}

\section{Related Work} \label{sec:relatedwork}

It has been shown in prior work that geolocation can be determined from images. % (e.g. examples in related work of \cite{yang2020protecting}). 
For example, PlaNet \cite{weyand2016planet}, an open-source deep learning approach pairing ground-level views and satellite data (see examples in Fig. \ref{fig:planet_results}), can predict a set of plausible locations of any image, including nature scenes. Most of these methods require multiple images for better performance, which are readily available for camera trap image collections. Other approaches focus on identifying objects in the scene, and using those identities to predict the locations \cite{workman2015wide}. Preventing locations from being revealed from images has also been considered in previous work, should this be necessary for camera trap images \cite{yang2020protecting}. In particular, \cite{yang2020protecting} applies to general image collections, such as those that might be posted to social media by users, and strategically deletes images until the location is ambiguous. However, all camera trap images taken from a single camera will have the same background, making it difficult to strategically remove images to reduce geolocalizability in a set of camera trap images. Camera trap images may also have very specific local landmarks, such as a well-known rock or tree, known to those familiar with an area but potentially hidden from generic deep learning methods. %, and those landmarks may be present in all images, making it impossible to strategically remove images to reduce geolocalizability in a set of camera trap imagery. 

% \section{What might make an image geolocalizable?}
% For images in cities, geolocalization is typically based on recognizable landmarks and geometries, such as relationships between buildings and roads, heights of buildings, and strong architectural features or signage. The more recognizable a feature (such as a famous landmark or horizon), the easier an image is to geolocate. If you see an image containing the Chrysler building, it's easy to know that you're most likely in NYC. If you can see the outline of Mount Rainier, you're most likely in or near Seattle. An image of the exterior of a nondescript chain hotel or stretch of highway might be more difficult to place.  The same might be said of camera trap imagery: if your image contains a noticeable landmark (for example a set of large rocky outcroppings or a body of water), it might be easier to estimate its location. In contrast, an image of dense undergrowth is only as potentially geolocalizable as your ability to recognize and model the distributions of its visible flora and fauna, and your ability to estimate a latitudinal band based on the timings of sunrise and sunset. One simple way to restrict the potential geolocalizability of your camera trap data could be to consciously place cameras in positions where the horizon and/or any known landmarks are not visible.

\section{Case study with Mpala Research Center}

%As a simple proof-of-concept to back up this intuition, we investigate our own ability to estimate the geolocation of a camera trap.
Our goal with this case study is to simply prove that geolocation is possible from camera trap imagery and metadata, indicating that sensitive animal locations could be vulnerable. We focus on % we investigate the concept that landmarks may inform geolocalizability in a case study in 
Mpala Research Center and explore both an off-the-shelf deep learning method for geolocalization, as well as a human-in-the-loop method.

\subsection{Mpala camera traps}
\begin{figure}
    \centering
    \includegraphics[width=0.9\linewidth]{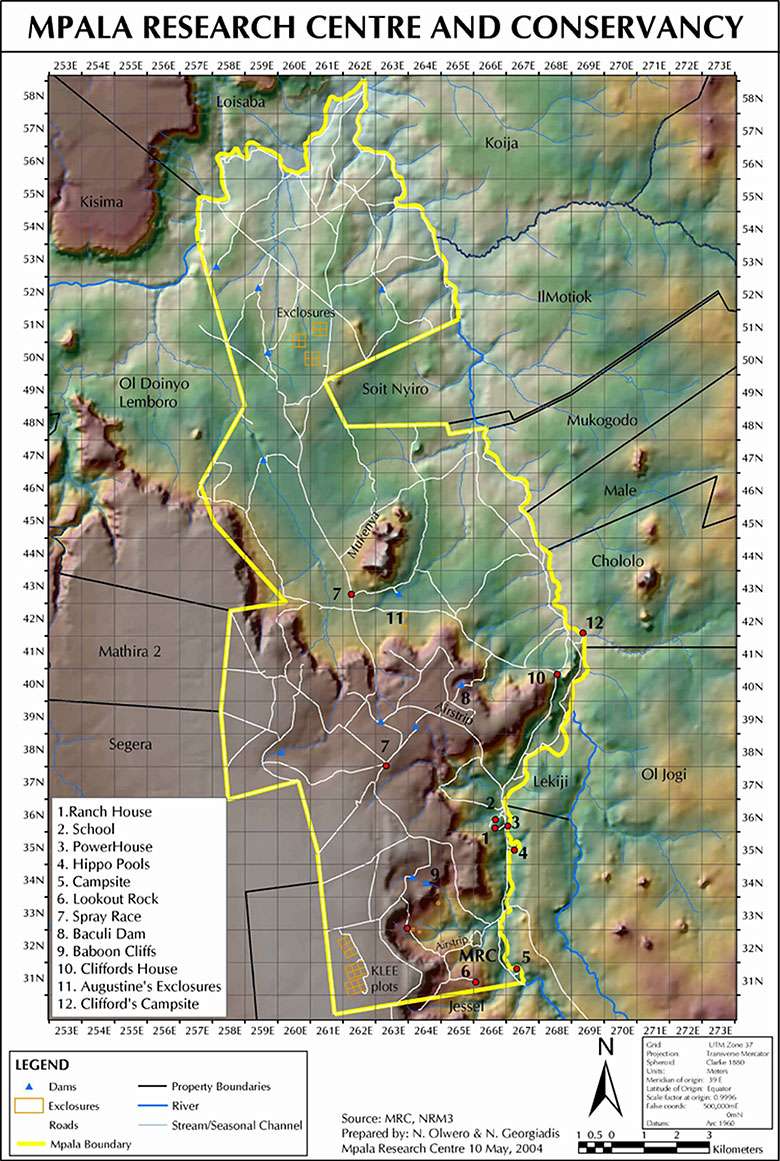}
    \caption{Map of Mpala Research Centre in Laikipia, Kenya, where the camera traps we studied were located. }%We used a network of 100 camera traps placed at Mpala as a preliminary testbed to investigate human-in-the-loop geolocalization from camera trap imagery.}
    \label{fig:mpala}
\end{figure}

The network of cameras we selected for our proof-of-concept is located at Mpala Research Center in Laikipia, Kenya. These 100 camera traps were initially placed as part of the 2020 Great Grevy's Rally, and have been continually collecting data over the past year. They capture a variety of habitats and backgrounds, including open savanna, two types of forested area, changes in elevation, and sites with visible horizon and without. You can see the diversity of landscapes across Mpala in the map in Fig. \ref{fig:mpala}.

\subsection{Off-the-shelf results}
\begin{figure}
    \centering
    \includegraphics[width=0.9\linewidth]{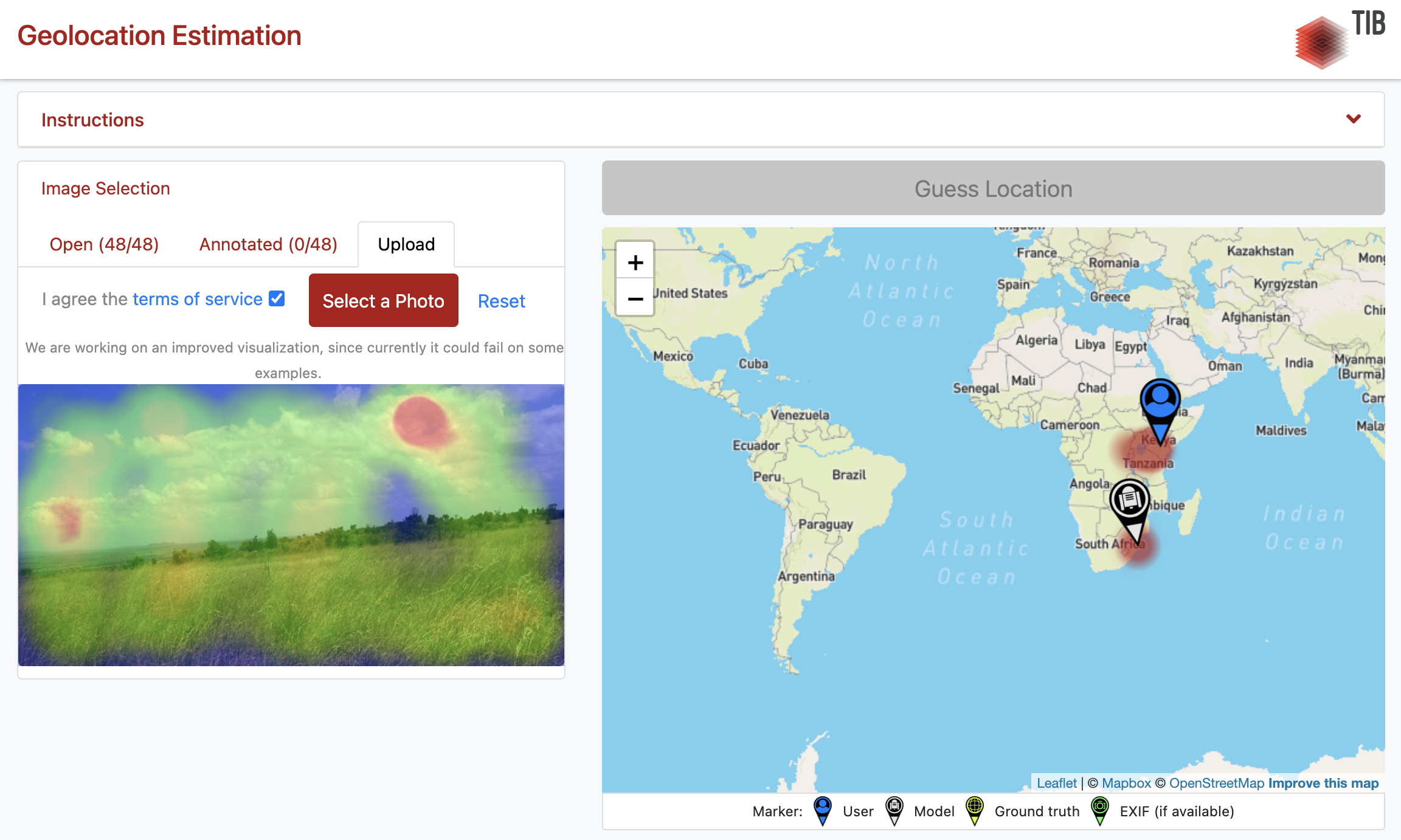}
    \caption{PlaNet results on an image from a Mpala camera trap. The blue marker is an approximation of the ground truth location, the white marker is the model prediction.}
    \label{fig:planet_results}
\end{figure}

We tried the existing PlaNet model on examples of camera trap data from Mpala, with results in Fig. \ref{fig:planet_results}. Given images from Mpala, PlaNet predicted large potential location areas covering Kenya, Tanzania, and South Africa. This may be due to the large amount of animal safari-based ecotourism in these countries. These areas are much larger than the potential randomness in location prescribed by most geo-obfuscation policies, rendering the off-the-shelf model unhelpful in further localizing the cameras. However, given camera trap-specific training data, similar deep learning-based methods may prove significantly more accurate. It is also interesting to note that the model seems to focus attention on the sky, which may imply that we should minimize the visibility of landmarks and horizons.

\subsection{Satellite Rasters}
% \smb{describe publicly available geo rasters}
We primarily utilized two sources of imagery from Google Earth Engine, specifically, (i) elevation data \cite{elevation}, which were collected in 2000 with a native resolution of 30 m and ranging from  about 1600-1800m in our region,  and (ii) Sentinel data \cite{sentinel}, specifically the red, green, and blue bands, which were collected in 2020 at a native resolution of 10 m. 
We downloaded each through Google Earth Engine's platform at 10 m resolution (the minimum of the two) over the same area to cover Mpala Research Centre, and all 100 camera traps. We then stacked these to form a multi-layer GeoTIFF. 

We also considered using a landcover map from Google Earth Engine \cite{buchhorn2020copernicus}, but we found that the classes did not have a great deal of distinction or resolution over our particular area of interest. We also note that if you know what part of the world you are in, and approximate sunrise and sunset directions, then it is possible to guess the approximate camera facing from just a few images (see Fig. \ref{fig:estimating_camera_facing}).  Methods for automatically determining sun direction \cite{wehrwein2015shadow} and camera position \cite{caulfield2004direction} based on shadows have been investigated in the computer vision literature, and could be used to scale up facing estimation for a large set of cameras. % Once you have an estimate for the facing direction of your camera, this can be used to further reduce the potential camera locations. We believe this could further reduce the search area reported in Table \ref{tab:results} in future work. 
These and other data could certainly be included in the future to further improve geolocation results.

\subsection{Human-in-the-loop geolocalization}

\begin{figure}

\centering
\subfloat[Elevation change (EC)]{\includegraphics[width=4cm]{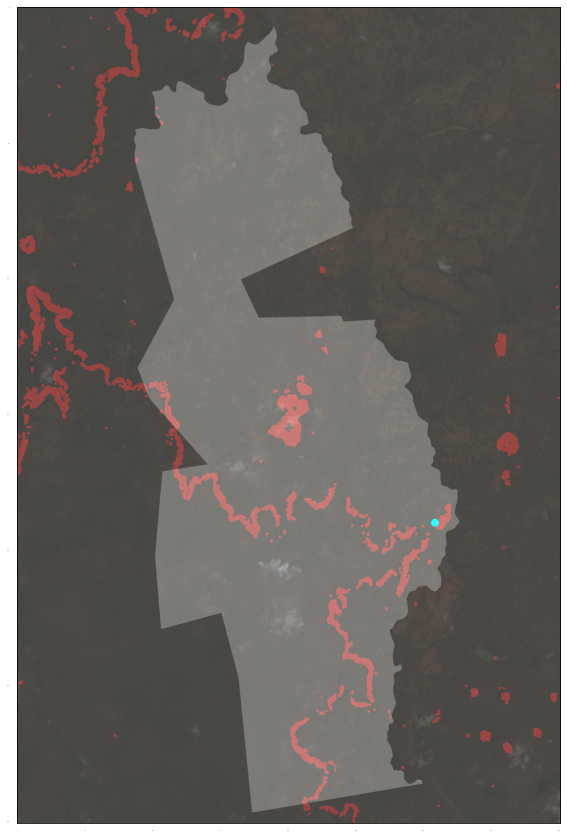}}
\subfloat[Red dirt (RD)]{\includegraphics[width=4cm]{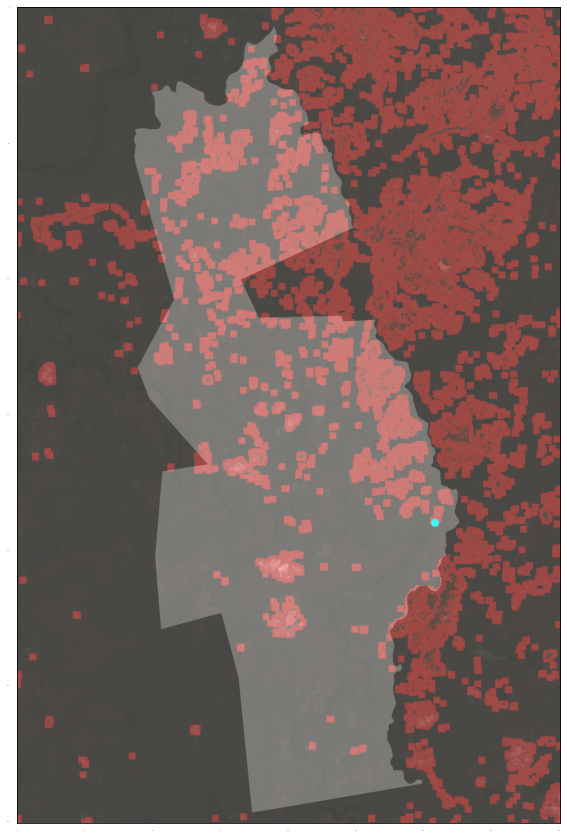}}\hfil
\subfloat[RD near EC]{\includegraphics[width=4cm]{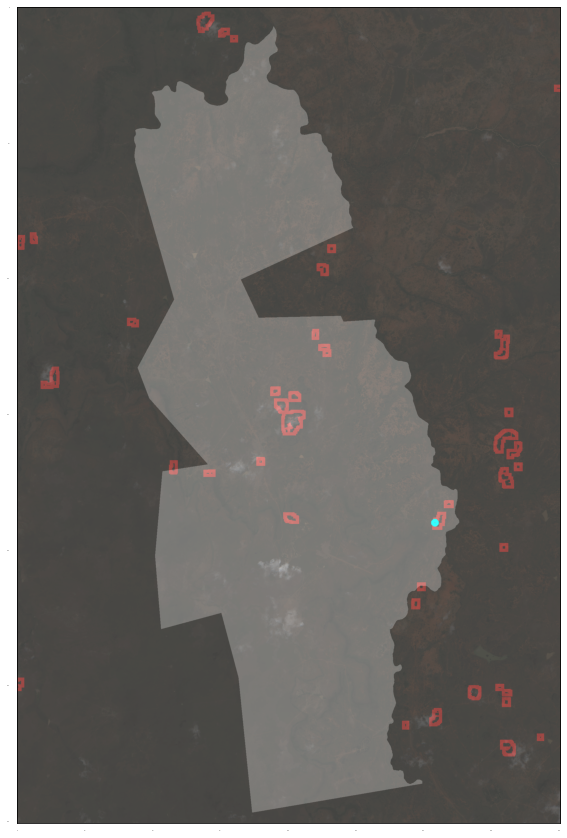}}  
\subfloat[RD near EC, within Mpala]{\includegraphics[width=4cm]{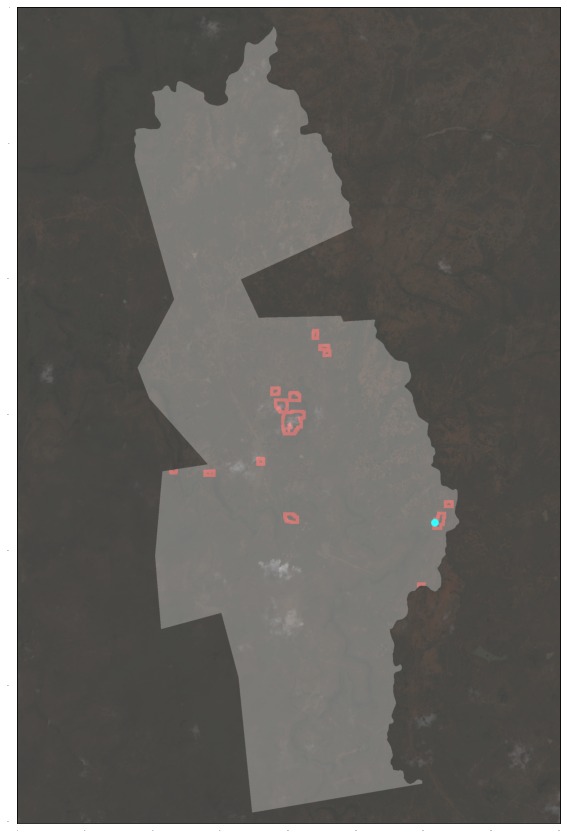}}\hfil
\subfloat[RD near EC, 10km obfuscation]{\includegraphics[width=4cm]{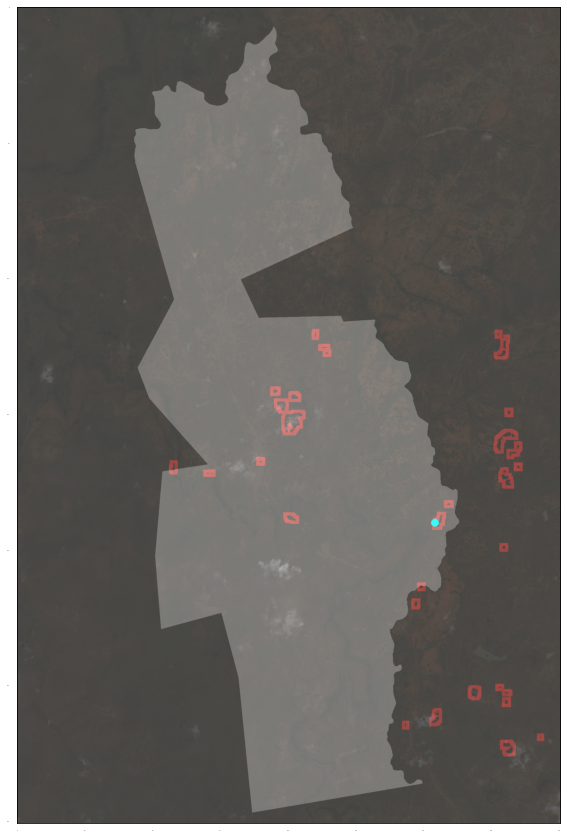}}
\subfloat[RD near EC, 1km obfuscation]{\includegraphics[width=4cm]{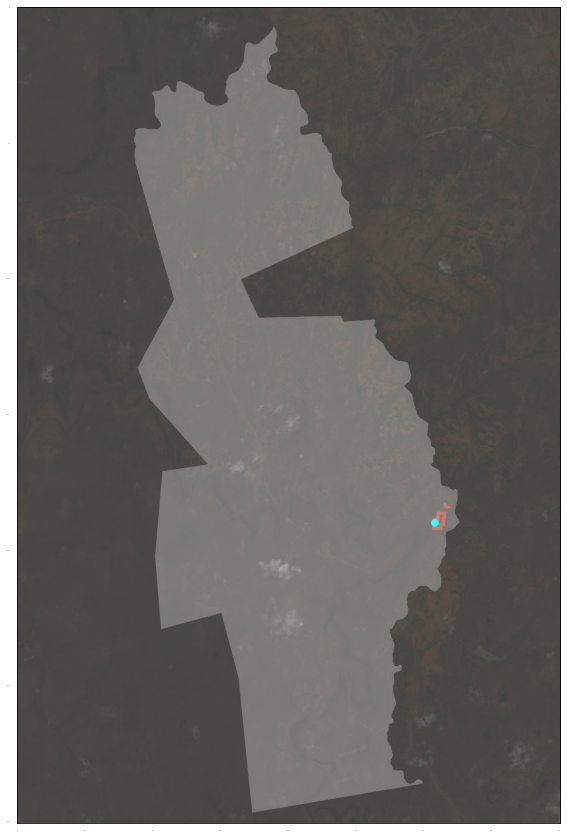}}
\caption{Human-in-the-loop geolocation filtering. In each example, the park boundary has been overlaid for context, and the camera location is represented by a small blue dot. The red portions of the image are  ``potential locations'' for the camera.} % More details on each filtering method can be found in the text.}
\label{figure}
\end{figure}

\begin{figure}
    \centering
    \includegraphics[width=0.95\linewidth]{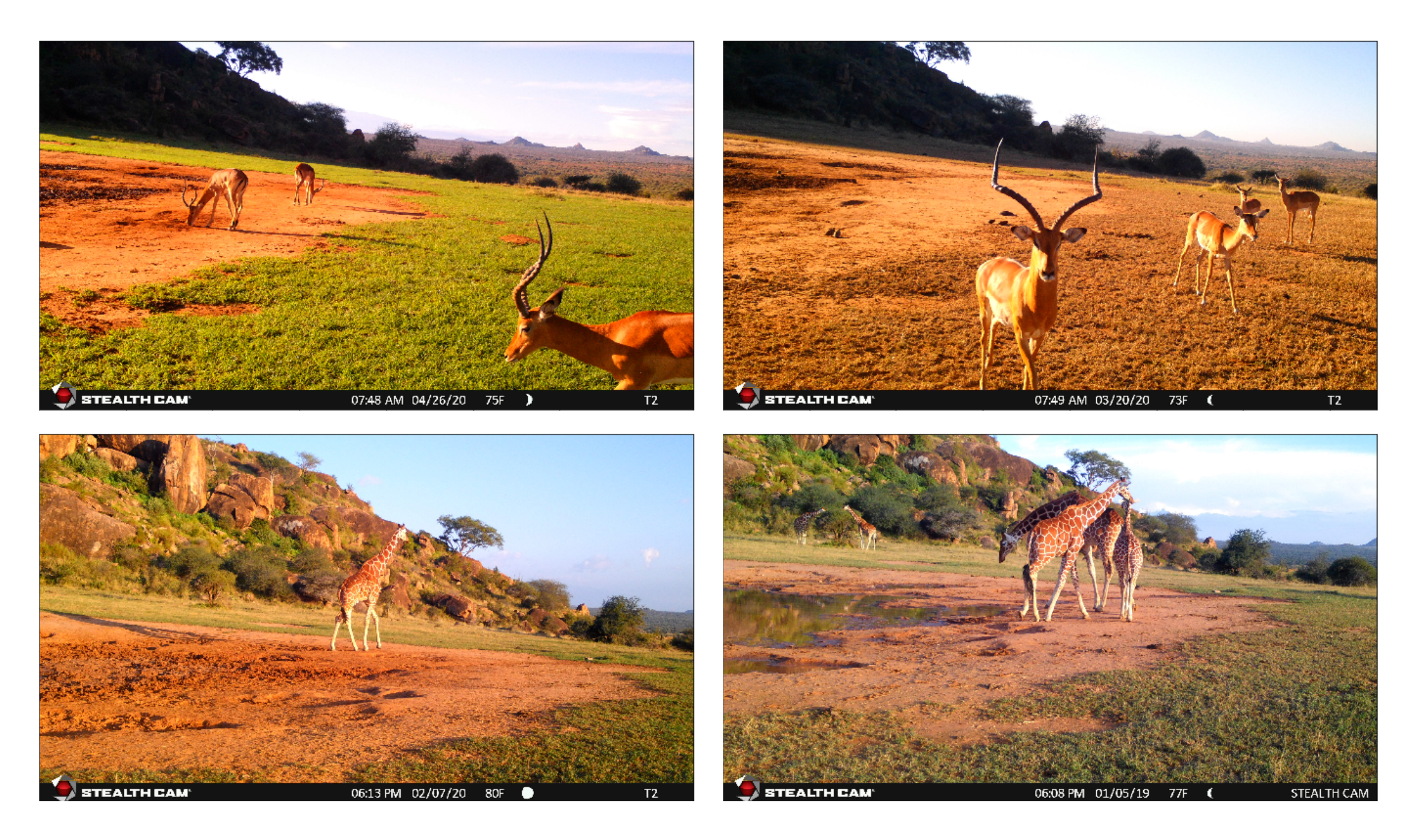}
    \caption{Images taken at sunrise (above) and sunset (below) imply that this camera is facing south by southeast.}
    \label{fig:estimating_camera_facing}
\end{figure}

We manually sampled one location from the camera traps to attempt to geolocate. We chose this location because it seemed to have recognizable features, for example, red-tinted soil and a large rock nearby (see examples in Fig. \ref{fig:estimating_camera_facing}), as we discussed in Sections \ref{sec:intro} and \ref{sec:relatedwork}. %We first manually examined the camera trap image for features, e.g., hills, rocks, trees, animals.
We decided to look for exactly these traits based on our observations. First, we computed the gradient of the elevation band and filtered for steep elevation change. We next searched for areas that were primarily red by thresholding the red band of the Sentinel data. We knew from our image that the camera trap was not in the red area itself, so we used morphological operators to select a small area surrounding red areas. In particular, we first did a closing operation to fill in gaps between small areas of the mask, then dilated this twice: first to represent a minimum distance away from the red area, then to represent a maximum distance away from the red area. We then subtracted the minimum dilation to get a ``donut'' shape around the red areas. We needed to do this for at least one of the two features in order to observe the areas of overlap (i.e., a red area nearby a rock). 
% \smb{probably change to algo format}

Once we found these two areas, we simply carried out an AND operation between the two masks. The remaining mask represented our candidate locations for the \emph{camera trap view}. However, the camera is placed at some distance to view this scene. Therefore, we carried out the same operation as when searching for ``near red'' areas. In particular, we estimated the distance of the camera from the landmarks in the image, and set the minimum and maximum dilation distances accordingly. We also adjusted the dilation distances to account for the fact that the camera might be located diagonally from the area of interest, and the kernels used for dilation are pixel- rather than distance-based, resulting in differing growth distance with each dilation diagonally vs. horizontally and vertically.

This provided us with final candidate locations for the camera trap. We therefore calculated the area of these locations by simply computing the final number of candidate pixels, and then multiplying by the area of these pixels, which is $10m * 10m = 100m^2$. This gives us the final ``searchable area,'' which we report in Table \ref{tab:results}, row 1.

\begin{table}[]
    \centering
    \begin{tabular}{|l|r|}
        \textbf{Filter type} & \textbf{Area (km$^2$)} \\
        RD near EC & 6.8301\\
        RD near EC, within Mpala & 2.0688 \\
        RD near EC, 10km obfuscation & 5.1373 \\
        RD near EC, 1km obfuscation & 0.2641 \\
    \end{tabular}
    \caption{Remaining area (in square kilometers) needed to search on foot to find the camera location, using our ``red dirt near elevation change'' human-in-the-loop filtering method and varying levels of geo-obfuscation.}
    \label{tab:results}
\end{table}

%\subsubsection{Estimating camera facing with shadows} \label{camera_pose}
%For instance, if you can see a landmark such as a hill roughly 200m away, and can approximate that your camera is facing southeast, then you know that you should be looking at distances 150-250m to the northwest side of areas of elevation gain.

\subsection{Geo-obfuscation}
%We synthesize varying the amount of geo-obfuscation and look at the effect it has on the geolocalizability of that camera. 
Mpala Research Centre is about 200 sq km in area, and the result from Table \ref{tab:results}, row 1 contains a slightly larger region due to the rectangular image encompassing the park. To synthesize the case where we release imagery and don't provide coordinates but do provide the park name, we restrict our final candidate locations by the boundaries of the park exactly. %, considering the case where we release imagery and don't provide coordinates but do provide the park name? 
Similarly, we repeat the calculation as though we were provided the coordinate geo-obfuscated by 10km and 1km. Our full results can be found in Table \ref{tab:results}. Providing the park name and using these simple image processing techniques can narrow the search space from 200 sq km to 2 sq km, and if the provided coordinates are known to be obfuscated by 1km this can narrow the search space to 0.26 sq km. For reference, 0.005 sq km is the area of an American football field, meaning 0.26 sq km is about 52 football fields. While this is still large, we believe that it would be possible to traverse this already, and likely further refine the predictions from satellite imagery with more sophisticated methods, including estimating camera facing and/or landcover, which could cut the search area in half. We emphasize that this reduction in search space was largely due to the presence of features in the image, especially the rock and soil landmarks. Again, this implies that we should further investigate avoiding or hiding such landmarks in camera trap imagery to protect the geolocation. %For example, by restricting the remaining area in the 1km obfuscated example to areas to the northwest of upward elevation change (based on our rough estimate of camera facing in Section \ref{camera_pose}), we could additionally filter the remaining area by half.

\section{Conclusions}
Using a very simple set of operations on human-generated heuristics based on publicly-available satellite rasters, we have shown that it is possible to drastically reduce the potential areas in which a camera may have been placed, meaning that poachers could theoretically find animals from public camera trap images. Based on our findings, one simple way to restrict the potential geolocalizability of your camera trap data could be to consciously place cameras in positions where the horizon and/or landmarks are not visible. %Though there is much to still investigate, our preliminary investigations show that geo-obfuscation might not be sufficient to protect the geoprivacy of a camera trap location. 
In future work, we hope to further analyze the performance of human-in-the-loop methods and investigate fully-automated methods for geolocalization based on deep learning to better understand how to protect sensitive species while promoting scientific understanding. We therefore bring this new challenge to the computer vision community: Can we analyze which types of features in a camera trap view lead to easier geolocalization? And if so, can we adversarially remove the localizeable features to preserve privacy without removing vital ecological information? %, such as differing levels of vegetation or the existance or lack of notable rocks or a visible horizon.

\section{Acknowledgements}

We would like to thank the entire team at Mpala Research Centre, and specifically Dan Rubenstein, Rosemary for their help managing the camera traps used to collect the data used in this case study. This work was supported, through funding, data storage, and computing resources, by Google AI for Nature and Society, Wildlife Insights, the Caltech Resnick Sustainability Institute, and NSFGRFP Grant No. 1745301, the views are those of the authors and do not necessarily reflect the views of these organizations.

{\small
\bibliographystyle{ieee_fullname}
\bibliography{main}
}

% \appendix
% \section{Dataset Statistics}

% Eli TODOs:
% \begin{itemize}
%     \item Characterize number of time points per site.
%     \item Characterize time range coverage at different sites.
%     \item Characterize cloud cover and fill pixels.
% \end{itemize}

% \section{QA Band Format}

\end{document}